\documentclass[runningheads]{llncs}
\usepackage{graphicx}
\usepackage{listings}

\usepackage{wrapfig}
\begin{document}
%

\title{Guiding Symbolic Natural Language \\ Grammar Induction \\ via Transformer-Based Sequence Probabilities}
\titlerunning{Guiding Grammar Induction via Transformers}
%

\author{Ben Goertzel\inst{1} 
\and Andr\'es Su\'arez-Madrigal\inst{1,2} 
\and Gino Yu\inst{2}}
%

%
\institute{SingularityNET Foundation, Amsterdam, The Netherlands
\url{https://singularitynet.io}\\
\email{ben@goertzel.org}
\and
The Hong Kong Polytechnic University, Kowloon, Hong Kong
\email{suarezandres@gmail.com}}
\maketitle              
\begin{abstract}
A novel approach to automated learning of syntactic rules governing natural languages is proposed, based on using probabilities assigned to sentences (and potentially longer word sequences) by transformer neural network language models to guide symbolic learning processes like clustering and rule induction.
This method exploits the learned linguistic knowledge in transformers, without any reference to their inner representations; hence, the technique is readily adaptable to the continuous appearance of more powerful language models.
We show a proof-of-concept example of our proposed technique, using it to guide unsupervised symbolic link-grammar induction methods drawn from our prior research.

\keywords{Unsupervised grammar induction \and Transformers \and BERT.}
\end{abstract}
%
%
%
\section{Introduction}

Unsupervised grammar induction -- learning the grammar rules of a language from a corpus of text or speech without any labeled examples (e.g. sentences annotated with human-created syntax parses) -- remains in essence an unsolved problem.
Although it has been approached for decades \cite{charniak1996statistical}, useful applications for restricted domains have been presented \cite{higuera2008introduction},
and state-of-the-art performance is improving \cite{li_dependency_2018}, the resulting grammars for natural language are still not able to properly capture its structure.

Bypassing explicit representations of the grammar rules, recent transformer neural network models have shown powerful abilities at language prediction and generation, indicating that at some level they internally ``understand'' those rules.
However, such rules don't seem to be found in the neural connections in these networks in any straightforward manner \cite{clark_what_2019,htut_attention_2019}, and are not easily extractable without supervision.
Supervised extraction of grammatical knowledge from the BERT \cite{devlin_bert_2019} network reveals that, to map the state of a transformer network when parsing a sentence into the sentence's parse, complex and tangled matrix transformations are needed \cite{hewitt2019structural}.

Here we explore an alternate approach: Don't try to milk the grammar out of the transformer network directly, rather use the transformer's language model as a {\it sequence probability oracle}, a tool for estimating the probabilities of word sequences; then use these sequence probability estimates to guide the behavior of symbolic learning algorithms performing grammar induction.
Our proposal is actually agnostic in the mechanism to find rules, and could synergize well with related efforts  \cite{schmid2011inductive,grave2015convex}; what we introduce is a novel and powerful way to guide the induction.
This is work in progress, but preliminary results have been obtained and look quite promising.

Full human-level AI language processing will clearly involve additional aspects not considered here, most critically the grounding of linguistic constructs in non-linguistic data \cite{tomasello2003}.  However,  the synergy between symbolic and sub-symbolic aspects of language modeling is a key aspect of generally intelligent language understanding and generation which has not been adequately captured so far, and we feel the work presented here makes significant progress in this direction.

\section{Methodology}

Transformer network models like BERT \cite{devlin_bert_2019}, GPT-2 \cite{radford2019language}, and their relatives provide probabilistic language models which can be used to assess the probability of a given sentence.   The probability of sentence $S$ according to such a language model tells you the odds that, if you sampled a random sentence from the model (used in a generative way), the output would be $S$.   If $S$ is not grammatical according to the grammar rules of the language modelled by the network, its probability will be very low.   If $S$ is grammatical but senseless, we assume from experimentation with these models, that its probability should also be quite low.

Having a sentence (or more generally word sequence) probability oracle of this nature for a language provides a way to assess the degree to which a given grammar $G$ models that language.   What one wants is that: the high-probability sentences according to the oracle tend to be grammatical according to $G$, the low-probability sentences according to the oracle are less likely to be grammatical according to $G$, and $G$ is as concise as possible.   The grammars that best fit these conjuncted factors are the best grammatical models of the language in question.

This concept could be used to cast grammar induction as a probabilistic programming problem, in a relatively straightforward but computationally exorbitant way.   Just sample random grammars from some reasonable distribution on grammar space, and evaluate their quality by the above factors.

What we propose here is conceptually similar but more feasible: Begin with a symbolic grammar learning algorithm which is capable of incrementally building up a complex grammar, then use sentence probability estimates from a neural language model to guide the grammar learning.  One could view this as an instance of the probabilistic programming approach, using a linguistic-theory-based heuristic method of sampling grammar space.

Our prior work on symbolic grammar induction \cite{glushchenko_programmatic_2019} uses two mains steps to build a dependency grammar from an unlabeled corpus. 
First, separate the vocabulary of interest into word categories (functionally equivalent to parts of speech, with a certain level of granularity). 
An implicit sub-step here is the disambiguation of polysemous words in the vocabulary, so that a single word could be assigned to more than one category.
Then, perform rule induction to find how words in these categories are connected to form grammatical sentences.

\begin{figure*}[]
\centering
\includegraphics[width=10cm]{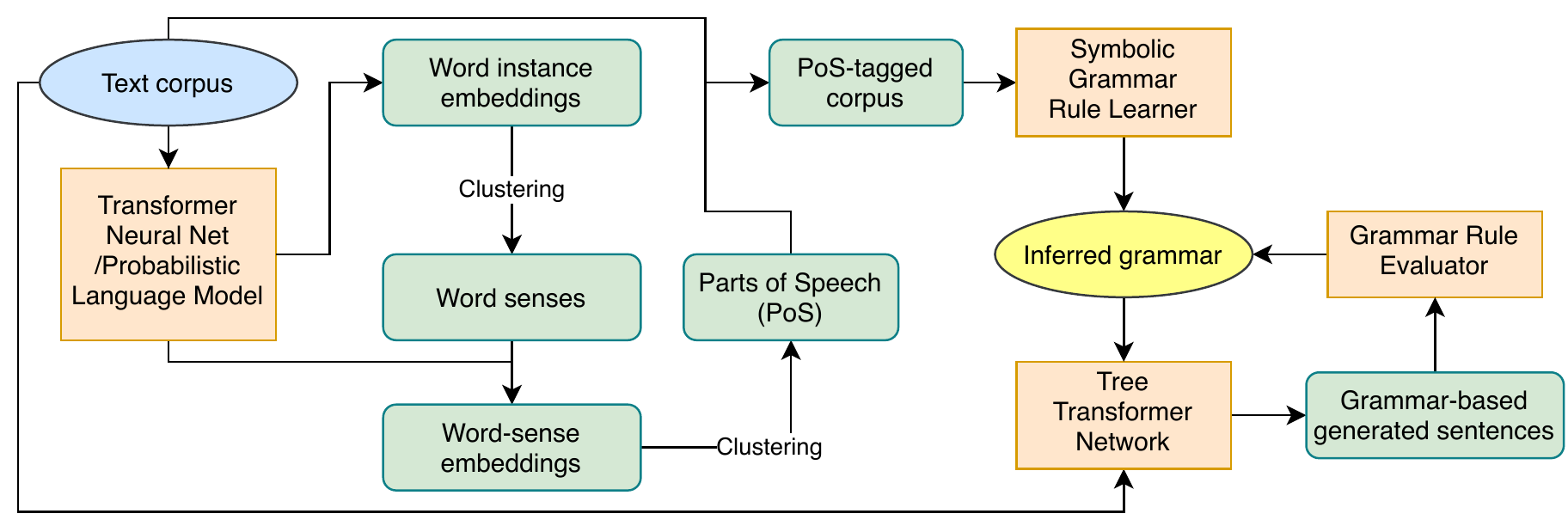}
\caption{High-level grammar learning architecture involving symbolic learning guided by estimated word sequence probabilities from a transformer network.}
\label{fig:neural-symbol-arch}
\end{figure*}

\noindent Our proposed approach, which enhances the aforementioned steps with the use of transformer language models, is depicted in Figure \ref{fig:neural-symbol-arch} and summarized as:

\begin{enumerate}
\item Infer word-senses and parts of speech from vectors built using a neural language model as a sentence probability oracle.
\item Infer grammatical rules from symbolic pattern-analysis of the corpus tagged with these senses and parts of speech.
\item Assemble a grammar incrementally from inferred rules. To evaluate whether a given rule should be included in the grammar:
\begin{itemize}
    \item Using a tree transformer network, generate a set of sentences consistent with the given rule, and others that follow mutations of the rule.
    \item Use a neural model as a sentence probability oracle to estimate whether the inferred rule leads to better generated sentences than its mutation(s).
\end{itemize}
\end{enumerate}

\noindent For our early experiments, we have chosen BERT \cite{devlin_bert_2019} as the transformer to use, but the idea could easily make use of similar unsupervised pre-trained networks.

\subsection{Assessing sentence probability}

\begin{figure*}[]
\centering
\includegraphics[width=6cm]{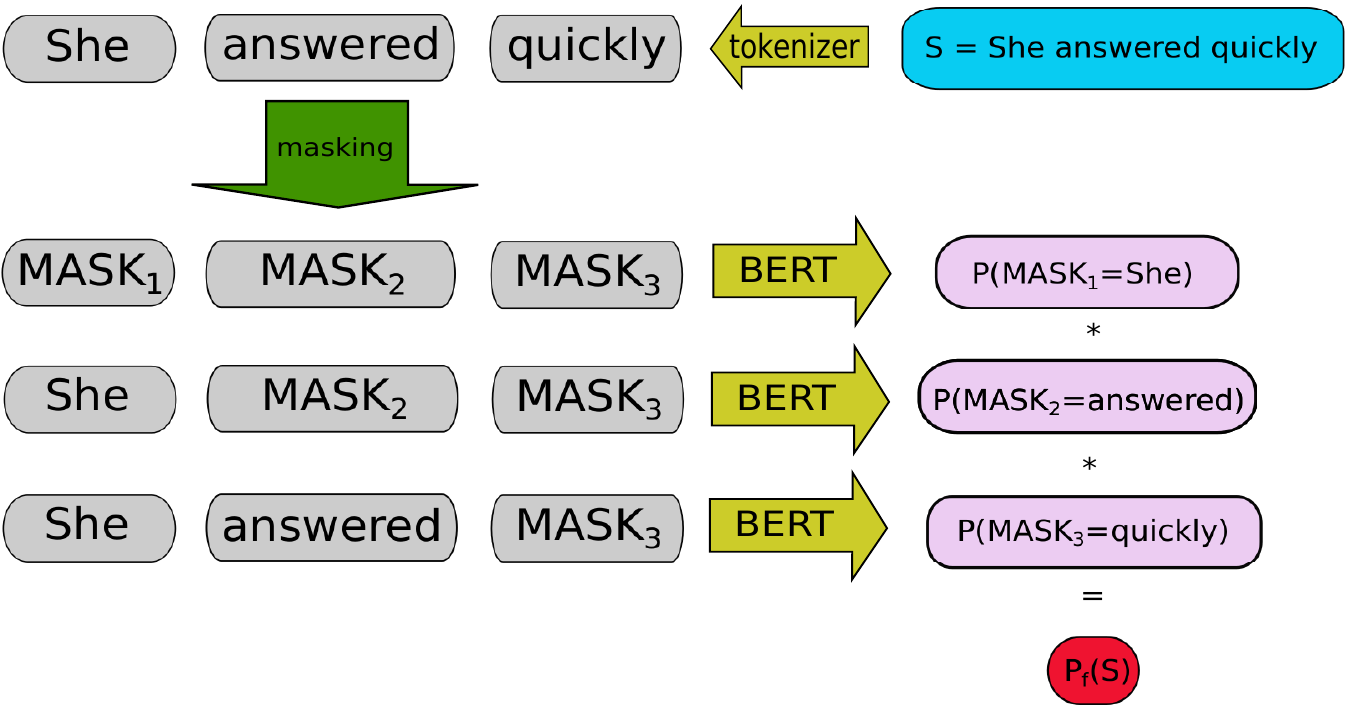}
\caption{Example of forward sentence probability calculation.}
\label{fig:prob_calc}
\end{figure*}

To explain details of our approach, we begin with the computation of sentence probability according to a neural language model (illustrated in Fig. \ref{fig:prob_calc}).   

Given a sentence $S = [w_0, w_1, ..., w_N]$, composed of N words $w_i, i \in [0, 1, ..., N]$, we want to calculate its probability $P(S)$.
A way to decompose that probability into conditional probabilities is:
\[
P_f(S) = P(w_0, w_1, ..., w_N) = P(w_0) \cdot P(w_1|w_0) \cdot P(w_2|w_0, w_1) \cdot ...
\cdot P(w_N|w_0, w_1, ..., w_{N-1}),
\]
which we call {\it forward sentence probability}.

A conditional probability $P(w_i|w_{i-1}, ..., w_0)$ can be obtained from BERT's masked word prediction model by taking the whole sentence, masking all the words which are not conditioned in the term (including $w_i$), and obtaining BERT's estimation for the probability of $w_i$.

To exemplify the idea, we summarize how to calculate the forward probability of the sentence ``She answered quickly''.
The probability is given by
\[
P_f(\mbox{She answered quickly}) = P(\mbox{She}) \cdot P(\mbox{She answered}|\mbox{She}) \cdot
P(\mbox{She answered quickly}|\mbox{She answered}).
\]
Each factor translates to a BERT Masked Language Model (MLM) prediction for a sentence with masked tokens.
For example, 
\[
P(\mbox{She answered}|\mbox{She}) = P(\mbox{MASK2=answered}|\mbox{She MASK2 MASK3}),
\]
and we get the probability that ``answered'' is predicted as the second token in the BERT MLM.

Now, to take advantage of BERT's bi-directional capabilities, we can estimate the sentence's {\it backwards probability} in a similar fashion:
\[
P_b(S) = P(w_0, w_1, ..., w_N) = P(w_N) \cdot P(w_{N-1}|w_N) \cdot P(w_{N-2}|w_{N-1}, w_N) \cdot ... \cdot P(w_0|w_1, w_2, ..., w_N)
\]

We finally approximate the sentence probability as the geometric-mean of the two directional ones:
\[
P(S) = \sqrt{P_f(S) \cdot P_b(S)}
\]


\subsection{Word Category Formation}

Following our prior work on symbolic grammar induction \cite{glushchenko_programmatic_2019}, and a number of previous works, we propose to generate embeddings for the words in the vocabulary and cluster them using a proximity metric in the embedding space.   Each final cluster can be considered a different word category, whose connection rules to other clusters will be defined in the induced grammar.  Unlike prior work, we use sentence probabilities as the embedding features.

\begin{figure*}[]
\centering
\includegraphics[width=12cm]{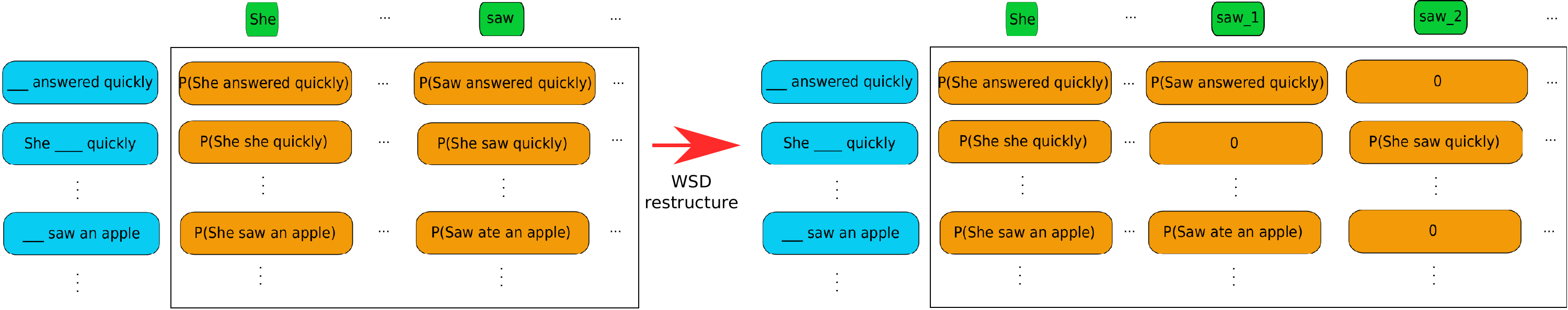}
\caption{{\it Left}: Matrix of words versus sentences-with-one-blank; each cell contains the probability of the given sentence filled with the given word.
{\it Right}: The matrix restructured after WSD.}
\label{fig:prob_matrix}
\end{figure*}

We expand each sentence in the corpus into $N$ sentences with a ``blank'' token in a different position, where $N$ is that sentence's length.
Each of those sentences with a blank is a feature for the word-vectors we will build.   Hence, we can think of a word-sentence matrix $M$, where rows are unique sentences with blanks in them, and columns are the words in the vocabulary (see Fig. \ref{fig:prob_matrix}).

We fill each cell in the matrix with the probability of the corresponding sentence-with-a-blank (row), when the blank is substituted by the corresponding word (column).
That is, if $S'_i$ is the sentence-with-a-blank in row $i$ and $w_j$ is the word in column $j$, then the cell $M_{i,j} = P(S'_i|\mbox{blank filled with }w_j)$.   

Once the matrix is filled, word categories are obtained by clustering the obtained word vectors (columns of the matrix).  Or, if one has performed word sense disambiguation (which can be done based on different computations from this same matrix, as will be described below), by clustering similar vectors corresponding to word senses.


\subsection{Word Sense Disambiguation}
\label{ssec:MethodWSD}

Word embeddings obtained from transformer networks by supervised learning have been used to disentangle word senses \cite{wiedemann_does_2019}; here we attempt this task in an unsupervised manner.  
From an unlabeled training corpus, we obtain a transformer embedding for each instance of each word in its given context.  Then, for each word in the vocabulary, we gather all of its embeddings and cluster them; we consider the resulting clusters as different word senses.

Specifically, a word-instance can be represented by a vector 
whose components are given by the probability that the neural language model assigns to the sentences (and discourse contexts) obtained by replacing such word instance with each word in the vocabulary.

Consider the word-instance ``test'' in ``The \underline{test} was a success''.
If the corpus vocabulary is $V = (\texttt{frog}, \texttt{which}, ...)$ then we can represent this instance's intension (contextual properties) with the vector $I$:

\begin{itemize}
\item I(test, The \underline{ }\underline{ }\underline{ }\underline{ } was a success)[1] = P(The \underline{frog} was a success)
\item I(test, The \underline{ }\underline{ }\underline{ }\underline{ } was a success)[2] = P(The \underline{which} was a success)
\item $\ldots$
\end{itemize}

\noindent Noticeably, the matrix obtained this way is the same one used for word-category formation; only, instead of performing clustering over the word vectors (columns), we need to independently cluster the rows that belong to instances of the same word to find their different senses.

\subsubsection{Word Category Formation in Depth.}
Once polysemy is taken care of, we can perform word-categorization over word-senses, allowing the same word to be assigned to different parts of speech (PoS) (e.g. ``test'' as a noun and as a ``verb'').
We need, however, to re-structure the sentence probability matrix to express word-senses as columns before grouping them into PoS.
This is done by reassigning the previously-calculated probabilities to the correct word-sense.

Starting from the original matrix $M$, we zero-initialize a disambiguated matrix $M'$ with the same number or rows, and as many columns as word-senses.
For a given entry in the original matrix, $M_{i,j}$, corresponding to sentence $S_i$ and vocabulary word $w_j$, we need to decide to which of its senses to assign it to.
If $w_j$ has only one sense, the decision is trivial; otherwise, we take the embedding for sentence $S_i$ (that is, the entire row $M_i$, as in the WSD process), and measure its distance from the centroids of the different senses for $w_j$ obtained in the WSD step.
The closest sense gets assigned the value $M_{i,j}$, and the rest keep a zero.
This way, we build word-sense embeddings by using the columns of $M'$; clustering these embeddings creates PoS categories and finer-grained syntactico-semantic categories.
Figure \ref{fig:prob_matrix} illustrates the disambiguated probability matrix.

 



\subsection{Grammar Induction}

After word categories are formed, grammar induction can take place by figuring out which groups of words are allowed to link with others in grammatical parses.  A grammar can be accumulated by starting with one rule and adding more incrementally, using the neural language model to evaluate the desirability of each proposed addition.  The choice of candidate rules is made by a symbolic rule induction algorithm; so far we have used the Grammar Learner process described in \cite{glushchenko_programmatic_2019}.

For a grammar rule proposed as an addition to the partial grammar already learned, we generate sentences that use that rule within the given grammar and obtain their sentence probabilities $P(S)$.  Then we corrupt the rule in some manner, adjust the grammar accordingly, generate sentences from this modified grammar starting with the mutated rule, and evaluate their $P(S)$.   If the sentences from the modified grammar decrease significantly in quality (where the threshold is a parameter), then the original rule is taken as valid.   The rationale is that correct grammar rules will produce better sentences than their distortions.

In the case of the link grammar formalism \cite{sleator1995parsing}, which we have used in our work so far, a grammar rule consists of a set of disjuncts of conjunctions of typed ``connectors'' pointing forward or backward in a sentence.   A mutation of this type of rule can be the swapping of each connector in the rule, which also implies a word-order change.

For example, if we have a rule $R$ that connects the word ``kids'' with the word ``the'' on the left and the word ``small'' also on the left, in that order:

\texttt{kids: small- \& the-},

\noindent which allows the string ``the small kids'', then the mutated rule $R^*$ would be

\texttt{kids: small+ \& the+},

\noindent which accepts the string ``kids small the''\footnote{Notice that connectors in the rules for small and kids also have to be modified to accommodate this mutation, i.e. they need to swap \texttt{kids+ to kids-}}.

\begin{figure*}[]
\centering
\includegraphics[width=7cm]{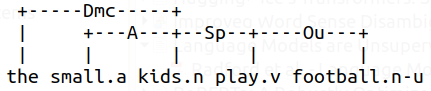}
\caption{Link-parse of "The small kids play football" according to the standard English link grammar dictionary \cite{sleator1995parsing}.}
\label{fig:link-parse}
\end{figure*}

This methodology requires a way to generate sentences from proposed grammars.   One approach is to use a given grammar to guide the attention within a Tree Transformer \cite{Wang2019}.  The standard Tree Transformer approach guides attention based on word-sequence segmentation that is driven by mutual information values between pairs of adjacent words.   One can replace these probabilities with mutual information values between pairs of words that are linked in partial parses that agree with a provided grammar.

Currently we are using a simpler stochastic sentence generation model in our proof-of-concept experiments, and planning to shift to a Tree Transformer approach for the next phase of work.

So, the rule $R$ guides the generation of sentences like $S =$ ``The small kids play football'' (see its Link-parse in Fig. \ref{fig:link-parse}).  The rule $R^*$ guides the generation of sentences like $S^* =$``Kids small the play football''.  The language model says $P(S) > P(S^*)$, thus arguing in favor of adding $R$ to one's grammar  (and then continuing the incremental learning process).

Alternatively, instead of producing mutated rules, one could also compare the probabilities of sentences generated with the rule under evaluation against those of a set of reference sentences of the same length, like those in the corpus used to derive the grammar, or the word categories obtained previously.

\section{Proof of concept (POC)}

Scalable implementation and testing of the ideas described above is work in progress; here we describe some basic examples we have explored so far, which validate the basic concepts (but do not yet provide a thorough demonstration).     We chose to perform our initial experiments using BERT\footnote{In particular, we use Huggingface's implementation of BERT, contained in their ``transformers'' package \cite{Wolf2019HuggingFacesTS} \url{https://huggingface.co/transformers}}, due to its popularity in several downstream tasks (e.g. word sense disambiguation \cite{wiedemann_does_2019}).

Following the workflow of the grammar induction process, we first show an example of word sense disambiguation, then one for word category formation, and finally grammar rule evaluation.

\subsection{Word sense disambiguation}

For an initial simple experiment, we created a small corpus of 16 sentences containing 146 words, out of which 8 are clearly ambiguous (to an English speaker).
Both syntactic and semantic ambiguities were included.
We generated embeddings for each word instance in the corpus, as described in section \ref{ssec:MethodWSD}.
Clustering was performed with spherical clustering methods from Spherecluster\footnote{https://github.com/jasonlaska/spherecluster} \cite{banerjee2005clustering}, as well as out-of-the-box DBSCAN and OPTICS models in Python's scikit-learn library with the cosine-distance metric.

We found that SphericalKMeans clustering did the best job at separating word senses in our test corpus.
Setting the number of clusters to two, the algorithm achieved an F1-score of 0.91.
As examples, the disambiguation for the word ``fat'', which was perfect, looks as follows:
 {\tt\begin{small}\begin{lstlisting}
Cluster #0 samples:
santiago became FAT after he got married .
there are many health risks associated with FAT .
the negative health effects of FAT last a long time .
Cluster #1 samples:
the FAT cat ate the last mouse quickly .
there is a FAT fly in the car with us .
\end{lstlisting}\end{small}}

The clustering for ``time'', on the other hand, placed one instance in the wrong category, and looks like this:
 {\tt\begin{small}\begin{lstlisting}
Cluster #0 samples:
i was born and raised in santiago de cuba , a long TIME ago .
my mouse stopped responding at the same TIME as the keyboard .
the negative health effects of fat last a long TIME .
Cluster #1 samples:
you will TIME the duration of the dress fitting session .
TIME will fly away quickly .
\end{lstlisting}\end{small}}

The disadvantage of using this straightforward implementation of SphericalKMeans is that one has to specify the number of clusters to use.
When requesting more clusters than there are senses for a word, the algorithm spreads instances with similar meanings to different clusters.
This is especially the case with words that we wouldn't consider ambiguous, like function words (we have sought to filter these by explicitly not disambiguating the top 10\% most frequent words in the corpus).
However, this may not be a terrible problem in our use case, as the word category formation algorithm will simply create more word-sense vectors per word, which then it could cluster together in the same word category.  Future experiments will involve alternatives that automatically estimate the number of clusters to use.

 \subsection{Word category formation}
Here, working with the same corpus as for WSD, we used the disambiguation results described above to build word vectors, thus allowing for words to be catalogued in more than one group.   Rather than SphericalKMeans, we found that OPTICS, a method that doesn't require a parameter for the number of clusters and can leave vectors uncategorized (shown as Cluster \#-1), offers remarkable quality in most formed clusters (\#0-14), with a good level of granularity.

{\tt\begin{small}\begin{lstlisting}
Cluster #-1: [fat, fat, ate, last, mouse, mouse, quickly, quickly,
 ., there, there, many, many, health, health, associated, with, 
 with, stopped, responding, same, time, as, will, fly, fly, negative, 
 of, a, a, long, in, us, tomorrow, she, she, was, was, wearing, 
 lovely, brown, brown, dress, attendees, did, not, properly, for, 
 occasion, became, after, got, married, ', ', s, deteriorated,
and, de, ,, ago, fitting, wasn, t, year, smith, protagonize, ]
Cluster #0: [the, my, his, ]
Cluster #1: [born, able, ]
Cluster #2: [raised, growing, bought, ]
Cluster #3: [cat, keyboard, car, session, feed, family, microsoft, ]
Cluster #4: [duration, episode, series, ]
Cluster #5: [are, is, ]
Cluster #6: [morning, night, ]
Cluster #7: [away, out, ]
Cluster #8: [they, he, i, you, ]
Cluster #9: [risks, effects, ]
Cluster #10: [at, to, ]
Cluster #11: [santiago, cuba, ]
Cluster #12: [time, will, long, ]
Cluster #13: [dress, and, ]
Cluster #14: [of, in, ]
\end{lstlisting}\end{small}}
 
 \noindent An evident problem with this result is that most of the words remain uncategorized (in Cluster \#-1).  Although we would expect the full iterative grammar learning algorithm we propose to be able to live with that and cluster some of the remaining words in the next pass, we will first try to fine-tune the procedure to alleviate this situation, as well as explore some other clustering algorithms.   At the same time, we predict that the results will improve when we use a larger number of features (instead of only 16 sentences for a total of 146 different features).   A very simple expansion of the vocabulary to cluster (not shown) already showed a similar number of more populated clusters.
 
\subsection{Grammar Rule Evaluation}
We show a simple use case for grammar rule evaluation, using the basic rule modification strategy proposed in the methodology: swapping the direction of the connectors that make up a rule, and comparing the sentences generated with and without the mutation.

For this experiment, we created a proof-of-concept grammar with 6 words divided in 6 categories: determiner, subject, verb, direct object, adjective, adverb.
Then, we assigned relationships among the classes.
Using a semi-random sentence generator, this grammar produces sentences like ``the small kids eat the small candy quickly.'' (that being the longest possible sentence in this grammar).

We then introduced some extra spurious rules to the grammar by hand.
From a total of 21 rules (15 correct ones vs. 6 spurious ones), the grammar can generate sentences like ``kids eat the the small candy kids eat candy the small quickly quickly.'', which clearly shows that the grammar is not correct anymore (this grammar has loops, so this is not even the longest sentence permitted by these simple modification).

Finally, we ran a first version of the grammar rule evaluator, to find out that all of the spurious rules were rejected, as well as three of the ``correct'' rules.

We notice that among the ``correct'' rules that were discarded, at least one:
 {\tt\begin{small}\begin{lstlisting}
eat: kids-,
\end{lstlisting}\end{small}}
\noindent generates sentences with no direct object, like ``the kids eat.''
This sentence, although valid, might not be very common for the BERT model, and thus obtain a low probability.

Similarly, the reverse of this rule, as modified by the evaluation algorithm:
 {\tt\begin{small}\begin{lstlisting}
eat: kids+,
\end{lstlisting}\end{small}}
\noindent generates sentences like ``eat the kids.'', which is also grammatically valid, and maybe as common as the previous case.
This is a sensible explanation for the rule's rejection.

\section{Conclusion and Future Work}

Our proof-of-concept experiments give intuitively strong indication of the viability of the methodology proposed for synergizing symbolic and sub-symbolic language modeling to achieve unsupervised grammar induction.   The next step is to create a scalable implementation of the approach and apply it to a large corpus, and assess the quality of the results.  If successful this will constitute significant progress both toward unsupervised grammar induction, and toward understanding how different types of intelligent subsystems can come together to more closely achieve human-like language understanding and generation.

%
%
\bibliographystyle{splncs04}
\bibliography{ULL_oracle.bib}

\begin{thebibliography}{10}
\providecommand{\url}[1]{\texttt{#1}}
\providecommand{\urlprefix}{URL }
\providecommand{\doi}[1]{https://doi.org/#1}

\bibitem{banerjee2005clustering}
Banerjee, A., Dhillon, I., Ghosh, J., Sra, S.: Clustering on the unit
  hypersphere using von mises-fisher distributions. Journal of Machine Learning
  Research  \textbf{6} (2005)

\bibitem{charniak1996statistical}
Charniak, E.: Statistical language learning. MIT press (1996)

\bibitem{clark_what_2019}
Clark, K., Khandelwal, U., Levy, O., Manning, C.D.: What {Does} {BERT} {Look}
  {At}? {An} {Analysis} of {BERT}'s {Attention}. arXiv:1906.04341 [cs]  (Jun
  2019)

\bibitem{devlin_bert_2019}
Devlin, J., Chang, M.W., Lee, K., Toutanova, K.: {BERT}: {Pre}-training of
  {Deep} {Bidirectional} {Transformers} for {Language} {Understanding}.
  arXiv:1810.04805 [cs]  (2019)

\bibitem{glushchenko_programmatic_2019}
Glushchenko, A., Suarez, A., Kolonin, A., Goertzel, B., Baskov, O.:
  Programmatic {Link} {Grammar} {Induction} for {Unsupervised} {Language}
  {Learning}. In: Artificial {General} {Intelligence}, vol. 11654, pp.
  111--120. Springer International Publishing (2019)

\bibitem{grave2015convex}
Grave, E., Elhadad, N.: A convex and feature-rich discriminative approach to
  dependency grammar induction. In: Proceedings of the 53rd Annual Meeting of
  the Association for Computational Linguistics. pp. 1375--1384 (2015)

\bibitem{hewitt2019structural}
Hewitt, J., Manning, C.D.: A structural probe for finding syntax in word
  representations. In: Proceedings of the 2019 Conference of the North American
  Chapter of the Association for Computational Linguistics. pp. 4129--4138
  (2019)

\bibitem{htut_attention_2019}
Htut, P.M., Phang, J., Bordia, S., Bowman, S.R.: Do {Attention} {Heads} in
  {BERT} {Track} {Syntactic} {Dependencies}? arXiv:1911.12246 [cs]  (Nov 2019)

\bibitem{higuera2008introduction}
de~La~Higuera, C., Oates, T., van Zaanen, M.: Introduction: Special issue on
  applications of grammatical inference. Applied Artificial Intelligence
  \textbf{22}(1-2), ~1--3 (2008)

\bibitem{li_dependency_2018}
Li, B., Cheng, J., Liu, Y., Keller, F.: Dependency {Grammar} {Induction} with a
  {Neural} {Variational} {Transition}-based {Parser}. arXiv:1811.05889 [cs]
  (Nov 2018)

\bibitem{radford2019language}
Radford, A., Wu, J., Child, R., Luan, D., Amodei, D., Sutskever, I.: Language
  models are unsupervised multitask learners. OpenAI Blog  \textbf{1}(8), ~9
  (2019)

\bibitem{schmid2011inductive}
Schmid, U., Kitzelmann, E.: Inductive rule learning on the knowledge level.
  Cognitive Systems Research  \textbf{12}(3-4),  237--248 (2011)

\bibitem{sleator1995parsing}
Sleator, D.D., Temperley, D.: Parsing english with a link grammar. arXiv:
  cmp-lg/9508004  (1995)

\bibitem{tomasello2003}
Tomasello, M.: Constructing a Language: A Usage-Based Theory of Language
  Acquisition. Harvard University Press (2003)

\bibitem{Wang2019}
Wang, Y.S., yi~Lee, H., Chen, Y.N.: Tree transformer: Integrating tree
  structures into self-attention. In: EMNLP/IJCNLP (2019)

\bibitem{wiedemann_does_2019}
Wiedemann, G., Remus, S., Chawla, A., Biemann, C.: Does {BERT} {Make} {Any}
  {Sense}? {Interpretable} {Word} {Sense} {Disambiguation} with
  {Contextualized} {Embeddings}. arXiv:1909.10430 [cs]  (Oct 2019)

\bibitem{Wolf2019HuggingFacesTS}
Wolf, T., Debut, L., Sanh, V., Chaumond, J., Delangue, C., Moi, A., Cistac, P.,
  Rault, T., Louf, R., Funtowicz, M., Brew, J.: Huggingface's transformers:
  State-of-the-art natural language processing. ArXiv  \textbf{abs/1910.03771}
  (2019)

\end{thebibliography}
\end{document}